\documentclass[acmtog]{acmart}

\acmSubmissionID{483}

\usepackage{booktabs}
\usepackage{amsmath}
\usepackage{soul}
\usepackage{microtype}
\usepackage{tabularx}
\usepackage{array}
\usepackage{enumitem}
\usepackage{comment}
\usepackage{wrapfig}
\usepackage{caption}
\newcolumntype{C}[1]{>{\centering\arraybackslash}m{#1}}
\newcolumntype{R}[1]{>{\raggedleft\arraybackslash}m{#1}}

\newcommand{\eg}{e.\,g.,\ }
\newcommand{\ie}{i.\,e.,\ }
\newcommand{\etal}{et~al.\ }

\newcommand{\refSec}[1]{Sec.~\ref{sec:#1}}
\newcommand{\refFig}[1]{Fig.~\ref{fig:#1}}
\newcommand{\refEq}[1]{Eq.~\ref{eq:#1}}
\newcommand{\refTbl}[1]{Tbl.~\ref{tbl:#1}}

\def\figurePath{}

\def\myfigure#1#2{\begin{figure}[t]\centering\includegraphics*[width = \linewidth]{\figurePath#1}\vspace{-.2cm}\caption{#2}\label{fig:#1}\end{figure}}

\def\mycfigure#1#2{\begin{figure*}[t]\centering\includegraphics*[clip, width = \linewidth]{\figurePath#1}\vspace{-.2cm}\caption{#2}\label{fig:#1}\end{figure*}}

\def\mysection#1#2{\section{#1}\label{sec:#2}}
\def\mysubsection#1#2{\subsection{#1}\label{sec:#2}}

\definecolor{fixedcolor}{rgb}{.1,.6,.1}
\definecolor{fixedncolor}{rgb}{.8,.2,.1}

\soulregister\ref7
\soulregister\cite7
\soulregister\refFig7
\soulregister\cite7
\soulregister\ref7
\soulregister\pageref7
\soulregister\shortcite7
\soulregister\eg0
\soulregister\ie0
\soulregister\etal0

\DeclareGraphicsExtensions{.png,.jpg,.pdf,.ai,.psd}

\acmJournal{TOG}
\acmYear{2018}
\acmVolume{37}
\acmNumber{6}
\acmMonth{11}
\acmArticle{235}

\setcopyright{rightsretained}

\acmDOI{10.1145/3272127.3275110}

\usepackage[ansinew]{inputenc}

\usepackage{graphicx}
\usepackage{amsmath}
\usepackage{xcolor}

\hypersetup{draft} 

\begin{document}
\title[Monte Carlo Convolution for Learning on Non-Uniformly Sampled Point Clouds]{Monte Carlo Convolution\\ for Learning on Non-Uniformly Sampled Point Clouds}

\author{Pedro Hermosilla}
\affiliation{%
  \institution{Ulm University}
  \country{Germany}}
\email{pedro-1.hermosilla-casajus@uni-ulm.de}
\author{Tobias Ritschel}
\affiliation{%
  \institution{University College London}
  \country{United Kingdom}}
\email{t.ritschel@ucl.ac.uk}
\author{Pere-Pau V\'azquez}
\email{pere.pau@cs.upc.edu}
\author{\`Alvar Vinacua}
\email{alvar@cs.upc.edu}
\affiliation{%
  \institution{Universitat Polit\`ecnica de Catalunya}
  \country{Spain}}
\author{Timo Ropinski}
\affiliation{%
  \institution{Ulm University}
  \country{Germany}}
\email{timo.ropinski@uni-ulm.de}


\begin{teaserfigure}
   \includegraphics[width=\textwidth]{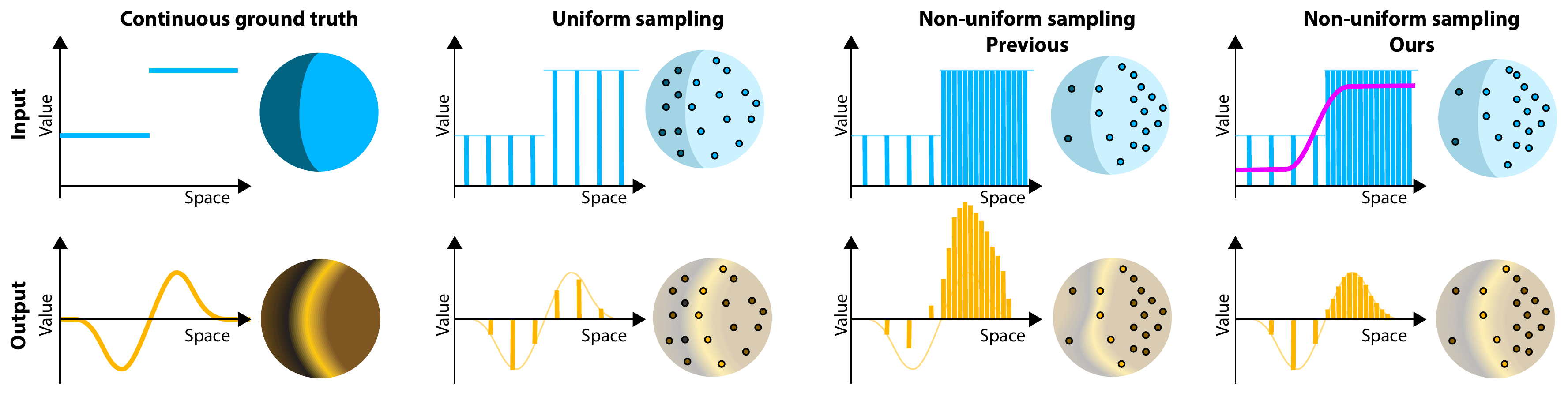}
   \vspace{-.4cm}
   \caption{
   Non-uniform sampling, which is inherent to most real-world point cloud datasets, has a severe impact on the convolved result signal. An input signal represented as 1D function and as projected on a 3D sphere \textbf{(top row)} is convolved with an edge-detection kernel to obtain the output signal represented as 1D function and as projected on a 3D sphere \textbf{(bottom row)}. The four columns illustrate the impact of the sample distribution on the convolution result. The ground truth continuous signal's filter response \textbf{(first column)} is faithfully captured when convolving uniformly sampled point clouds \textbf{(second column)}. In the case of non-uniformly sampled point clouds, state-of-the-art convolutional methods severely deviate from the desired filter response \textbf{(third column)}. With our interpretation of non-uniform convolution as a Monte Carlo estimate in respect to a given sample density distribution (illustrated by the pink line), we can compensate this deviation and obtain a filter response faithfully capturing that of the ground truth \textbf{(fourth column)}.}
   \label{fig:Teaser}
\end{teaserfigure}

\begin{abstract}
Deep learning systems extensively use convolution operations to process input data. Though convolution is clearly defined for structured data such as 2D images or 3D volumes, this is not true for other data types such as sparse point clouds. Previous techniques have developed approximations to convolutions for restricted conditions. Unfortunately, their applicability is limited and cannot be used for general point clouds. We propose an efficient and effective method to learn convolutions for non-uniformly sampled point clouds, as they are obtained with modern acquisition techniques. Learning is enabled by four key novelties: first, representing the convolution kernel itself as a multilayer perceptron; second, phrasing convolution as a Monte Carlo integration problem, third, using this notion to combine information from multiple samplings at different levels; and fourth using Poisson disk sampling as a scalable means of hierarchical point cloud learning. The key idea across all these contributions is to guarantee adequate consideration of the underlying non-uniform sample distribution function from a Monte Carlo perspective. To make the proposed concepts applicable to real-world tasks, we furthermore propose an efficient implementation which significantly reduces the GPU memory required during the training process. By employing our method in hierarchical network architectures we can outperform most of the state-of-the-art networks on established point cloud segmentation, classification and normal estimation benchmarks. Furthermore, in contrast to most existing approaches, we also demonstrate the robustness of our method with respect to sampling variations, even when training with uniformly sampled data only. To support the direct application of these concepts, we provide a ready-to-use TensorFlow implementation of these layers at \url{https://github.com/viscom-ulm/MCCNN}.
\end{abstract}

\maketitle



\section{Introduction}
While convolutional neural networks have achieved unprecedented performance when learning on structured data~\cite{Kaiming2015, Huang2016,Wang2017}, their application to unstructured data such as point clouds is still fairly new. Early methods have looked into fully-connected approaches and striven for permutation and rotational invariance~\cite{qi2017,qi2017plusplus,guerrero2016}. 
Unfortunately, a non-uniform sampling typically associated with real-world point cloud data, such as for instance resulting from the projective effect of a LiDAR scan, has not been a special focus of previous research. 
As the underlying sampling results in severe implications, as illustrated in~\refFig{Teaser}, we propose a new approach which has been developed with a special focus on non-uniformly sampled point clouds, achieving at the same time competitive performance on uniformly sampled data.

To make progress towards our goals, we take into account the original definition of convolution: an integral in an unstructured setting. By using Monte Carlo (MC) estimation of this integral, we will show that proper handling of the underlying sampling density is crucial, and will produce results surpassing the state-of-the-art.
To this end, we make four key contributions.

First, we represent the convolution kernel itself as a multilayer perceptron (MLP). Since a convolution kernel maps a spatial offset (Laplacian) to a scalar weight, representing and learning this mapping through an MLP is a natural choice. 

Second, we suggest using Monte Carlo integration to compute convolutions on unstructured data. Key is the adequate handling of the non-uniformity and varying-density of points from a Monte Carlo point of view. Averaging weighted pairs of points formally means to sample an MC estimate of an integrand. While MC requires to divide by the probability of each sample, this can be neglected in a uniform setting, as it would result in a division by a constant. However, when dealing with a varying sample density as present in non-uniformly sampled point clouds, failing to perform this normalization leads to a bias and consequently reduced learning ability. Therefore, we claim that our approach provides a new form of robust \emph{sampling invariance}, where for instance simply duplicating a point will not change the estimated integrand. Stating convolution as MC integration allows us to tap into the rich machinery of MC including (quasi) randomization~\cite{Niederreiter1992} and importance sampling~\cite{Kahn1953}. Consequently, the convolution becomes invariant under point re-orderings and typically works on receptive fields with a variable number of neighboring points.

Third, we show how this allows generalizing convolutions which use a single sampling pattern to convolutions that map from one sampling pattern to a different one with a higher or lower resolution. This can be used to learn (transposed) convolutions that change the level-of-detail for pooling or up-sampling operations. Even more general, we introduce convolutions that map from multiple input samplings to the desired output sampling allowing to learn combining information from multiple scales.

Fourth, we introduce Poisson disk sampling~\cite{Cook1986,Wei2008} as a means to construct a point hierarchy. It has favorable scalability compared to the state-of-the-art Farthest Point \cite{eldar1997} sampling and allows to bound the maximal number of samples in a receptive field.

The usefulness of these novelties is demonstrated by comparing our approach to state-of-the-art point cloud learning techniques for segmentation, classification, and normal estimation tasks. We will show that we outperform the state-of-the-art when learning on non-uniform point clouds, while we still achieve state-of-the-art performance for uniformly sampled point clouds.

\mysection{Previous Work}{PreviousWork}

A straight-forward method to enable learning on point clouds is to resample them to a regular grid and then applying learning approaches originally developed for structured data. While extensions to multiple resolutions exist, \eg based on octrees~\cite{Wang2017}, in this section, we will solely focus on those techniques which enable learning directly on unstructured data.

PointNet~\cite{qi2017} pioneered deep learning on unstructured datasets. It used a fully-connected network together with a clever machinery to achieve rotation and permutation-invariance. PointNet++~\cite{qi2017plusplus} added extensions to support localized sub-networks, but was not yet fully convolutional. PCP~Net~\cite{guerrero2016} allowed the inference of local properties like curvature or normals but was also not convolutional.

Klokov and Lempitsky \shortcite{klokov2017escape} presented a convolutional learner which used a $k$-d tree. However, since the leaf nodes of the tree had a fixed number of points, it was sensitive to varying density, being tight to the logic of building and querying $k$-d trees. In contrast, our approach uses a regular grid to access neighbors in constant time and thus works on multiple scales.

Shen~\etal\shortcite{Shen2017} also provided a translation-invariant but non-convolutional method, where convolution was replaced with the correlation of local neighborhood graphs and learned graph templates. While showing good performance on small problems, it remained particularly sensitive to the underlying graph structure.

Dynamic Graph CNNs by Wang~\shortcite{Wang2018} were convolutional. They employed a general notion of learnable operations on edges of a graph of neighboring points. In contrast to other approaches, they changed neighborhoods during learning, making the approach slightly more complex to implement and less efficient on large point clouds.

PointCNN~\cite{Li2018} was also convolutional, working on the $k$ nearest neighbors. They
used an MLP on the entire neighborhood to learn a transformation matrix which was used later to weight and permute the input features of the neighboring points. Then, a standard image convolution was applied to the transformed features.

SPLATNet~\cite{Su2018} sought inspiration from the permutohedral lattice~\cite{adams2010fast} where convolutions can efficiently be performed on sparse data in high dimensions while the filter kernels are discrete masks in lattice space. Uneven sample distributions in the lattice are addressed by ``convolving the 1'' \cite{adams2010fast} \ie repeating the convolution on a unit signal. Our work achieves the same, but without the complication of creating a lattice. 

Atzmon~\etal\shortcite{Atzmon2018} use radial basis functions defined on a discrete set of points to represent convolution kernels. The work is computationally demanding but provides invariance under global uniform resampling by construction. However, non-uniform sampling settings are not considered.






In concurrent work, SpiderCNN~\cite{Xu2018} used step functions to represent convolutions. The authors mentioned using MLPs, as we do as well, but found them to perform worse than step functions. Nevertheless, in our architectures, we found MLPs to perform well. We acknowledge that further work shall explore different continuous representations to parametrize learned convolutions.

Groh~\etal\shortcite{Groh2018} suggested using a linear function as a representation of an unstructured convolution kernel. With considerations how dissimilar 2D convolution kernels are from linear functions, we think an MLP to describe a kernel is worth exploring. Thanks to the simplicity we share in our approach, they demonstrated excellent scalability to millions of points, but the problem of non-uniform sampling is not touched upon.

\myfigure{KNNIssue}{A $k=2$-nearest neighbor receptive field \textbf{(blue circle)} in a scene with non-uniform sampling of the same house-like geometry changes scale.}

Most of the existing methods are based on convolving the $k$ nearest neighbors (using a hierarchical structure or not)~\cite{klokov2017escape, Shen2017, Wang2018, Li2018, Xu2018, Groh2018}. 
This approach is not robust in non-uniform sampling settings, since adding more points into a region in the space will reduce the $k$ nearest neighbors to a small volume around each point, capturing different features from those the kernel was trained on, i. e. it shrinks in densely populated areas and grows in sparse ones.
We see in \refFig{KNNIssue} how this would prevent constructing a sampling-invariant ``house detector''. 
Aztmon~\etal\shortcite{Atzmon2018}, although not considering non-uniformity in their paper, transform the point cloud to volumetric functions which can be robust to non-uniformly sampled point clouds. However, some computations of their method are quadratic on the number of points which makes their method not scalable. PointNet++~\cite{qi2017plusplus}, on the other hand, computes features locally, which makes it scalable. Moreover, it was tested with non-uniformly sampled point clouds by simulating the properties of LIDAR scans. Nevertheless, their method selects a fixed number of random samples within a radius around the points and thus does not consider the point density in its computations, which, as we will demonstrate, can lead to errors.

\mysection{Convolution kernels}{MLPKernels}
Here, we will first recall the definition of a convolution (\refSec{Convolution}).
Then, we will introduce a kernel representation using MLPs (\refSec{MLPDetails}) which shall allow efficient and simple learning on irregular data.

\mysubsection{Convolution as an integral}{Convolution}
Recall the definition of a convolution as an integral of a product:

\begin{equation}
(f\ast g)(\mathbf x)
=
\int
f(\mathbf y)
g(\mathbf x-\mathbf y)
d\mathbf y
\label{eq:conv}
\end{equation}

\noindent where $f$ is a scalar function on $\mathbb R^3$ to be convolved and $g$ is the \emph{convolution kernel}, a scalar function on $\mathbb R^3$. In our particular case, $f$ is the \emph{feature function} for which we have a set $\mathcal S$ of discrete samples $\mathbf x_i\in\mathcal S$ (our data points). If for each point no other information is provided besides its spatial coordinates, $f$ represents the binary function which evaluates to $1$ at the sampled surface and $0$ otherwise. However, $f$ can represent any type of input information such as color, normals, etc. For internal convolutions, \ie those which are subsequent to the input layer, it can also represent features from a previous convolution.

\paragraph*{Translation-invariance.} As the value of $g$ only depends on relative positions, convolution is translation-invariant.

\paragraph*{Scale-invariance.} Since evaluating the integral in \refEq{conv} over the entire domain can be prohibitive for large datasets, we limit the domain of $g$ to a sphere centered at $0$ and radius $1$. In order to support multiple radii, we normalize the input of $g$ dividing it by the receptive field $r$. In particular, we choose $r$ to be a fraction of the scene bounding box diameter $b$, for instance $r=.01 \cdot b$. Doing so results in scale-invariance. Note, that this construction results in compactly-supported kernels that are fast to evaluate.

\paragraph*{Rotation-invariance.} Note, that we do not achieve and seek to achieve rotation-invariance. Typical image convolutions are not rotation invariant either and succeed nonetheless.

\myfigure{GMLP}{
Evolution of MLP kernels:
\textbf{a)} A na\"ive solution would map one 2D offset $x,y$ (\textbf{blue dots}) to one scalar result $g$ (\textbf{orange dots}).
\textbf{b)} We suggest extending this to multiple outputs, \eg $g_1$ and $g_2$, which speeds up computation and reduces the number of learnable parameters.
\textbf{c)} 
Our 3D implementation uses two hidden layers of 8 neurons outputting 8 kernel values.
In this example we require $2\times 8\times 8+3\times 8=152$ operations to learn and compute while a na\"ive approach needs $8\times(3\times 8+8\times 8+8)=768$.
}
   
\mycfigure{MCIntegration}{
   Steps of our MC convolution.
   For a given point $\mathbf x$ \textbf{(a)} the neighbors within the receptive field $r$ are retrieved \textbf{(b)} to be used as Monte Carlo integration samples \textbf{(c)}.
   For each neighboring point $\mathbf y_j$, its probability density function, $p(\mathbf y_j|\mathbf x)$, is computed using \emph{Kernel Density Estimation}~\cite{rosenblatt1956, parzen1962} \textbf{(d)}.
   Based on the bandwidth used (pink disk), the neighboring points have different effects on the computation of $p(\mathbf y_j|\mathbf x)$ (pink gradient).
}

\mysubsection{Multilayer perceptron kernels}{MLPDetails}
We suggest to represent the kernel $g$, by a multilayer perceptron.

\paragraph{Definition}
The multilayer perceptron (MLP) takes as input the spatial offset $\delta=(\mathbf x-\mathbf y)/r$ comprising of three coordinates, normalized dividing them by the receptive field $r$. The output of the MLP is a single scalar. To balance accuracy and performance, we use two hidden layers of $8$ neurons each (see ~\refSec{Evaluation} for more details). We denote the hidden parameters as a vector $\omega$.

For layers with a high number of input and output features, the number of kernels and therefore the number of parameters the network has to learn is too high ($\#inputs \times \#outputs$). To address this problem, we use the same MLP to output $8$ different $g$'s, thus reducing the number of MLP's by a factor of $8$. \refFig{GMLP} presents an illustration of such an MLP, which takes three coordinates as input and outputs $8$ different $g$'s.

\paragraph{Back-propagation}
For back-propagation~\cite{rumelhart1986learning}, the derivative of a convolution with respect to the parameter $\omega_l$ of the MLP is

\begin{equation}
\frac{\delta f\ast g}{\delta \omega_l}
=
\int f(\mathbf y)
\frac{
\delta g(\mathbf x-\mathbf y)
}{
\delta \omega_l
}
d\mathbf y.
\label{eq:diff-conv}
\end{equation}

\mysubsection{Single- and multi-feature convolution}{SigleAndMultipleFeatures}
Our convolution consumes $M$ input feature functions and outputs $L$ convolved feature functions. Based on the way the convolved feature functions are calculated, we define two types of layers: \emph{Single-feature} spatial convolution and \emph{Multi-feature} spatial convolution. Single-feature spatial convolution outputs a scalar feature by convolving a scalar input feature. Therefore, in these layers, the number of output features is equal to the number of input features, $M = L$, and the number of kernels $g$ is also equal to $M$. The multi-feature convolution, on the contrary, is similar to the layers used in standard convolutional neural networks where each output feature is computed as the sum of all input feature functions convolved:
\begin{equation}
f_o
=
\sum_{i=0}^{M}
f_i\ast g_{o,i}
\end{equation}

\noindent These layers are more computationally demanding since they have to learn $M \times L$ convolution kernels $g$.

\mysection{Monte Carlo Convolution}{MCConvolution}
In this section, we will show how convolutions can be stated as a Monte Carlo estimate by relying on a sample's density function, which ultimately makes learning robust to non-uniform sample distributions.

\mysubsection{Monte Carlo integration}{MCIntegration}
In order to compute the convolution in each sample point, we have to evaluate the integral of Equation~\ref{eq:conv}. Since we only have a set of samples of our function $f$, we propose to compute this integral by using MC integration~\cite{Kalos1986}, which uses a set of random samples to compute the value of an integral.

\paragraph{Definition}
In our case, these samples comprise of the input data points or a (quasi) random subset. Therefore, an estimate of the convolution for a point $\mathbf x$ is

\begin{equation}
(f\ast g)
(\mathbf x)
\approx
\frac{1}{|\mathcal N(\mathbf x)|}
\sum_{j\in\mathcal N(\mathbf x)}
\frac{
f(\mathbf y_j)g\left(\frac{\mathbf x - \mathbf y_j}{r}\right)
}{
p(\mathbf y_j|\mathbf x)
},
\label{eq:MCConvPDF}
\end{equation}

\noindent where $\mathcal N(\mathbf x)$ is the set of neighborhood indices in a sphere of radius $r$ (the receptive field), and $p(\mathbf y_j|\mathbf x)$ is the value of the \emph{Probability Density Function} (PDF) at point $\mathbf y_j$ when the point $\mathbf x$ is fixed, \ie the convolution is computed at $\mathbf x$. \refFig{MCIntegration} provides an illustration of this computation.

Please note that, since our input data points are non-uniformly distributed, each point $\mathbf y_j$ will have a different value for $p(\mathbf y_j|\mathbf x)$.
It is also worth noticing, that the PDF depends not only on the sample position $\mathbf y_j$ but also on $\mathbf x$: How likely it is to draw a point does not only depend on the point itself, it also depends on how likely the others in the receptive field $r$ are.

Please finally note that, here and in the following, $\mathbf x$ is an arbitrary output point that might not be from the set of all input points $\mathbf y_j$. This property will later allow re-sampling to other levels or other irregular or regular domains. 

\paragraph{Back-propagation}
Back-propagation~\cite{rumelhart1986learning} with respect to the MLP parameters $\omega_l$ can also be estimated using MC:

\begin{equation}
\left(
\frac{
\delta f\ast g
}{
\delta \omega_l
}
\right)
(\mathbf x)
= 
\frac{1}{|\mathcal N(\mathbf x)|}
\sum_{j\in\mathcal N(\mathbf x)}
\frac{f(\mathbf y_j)}{p(\mathbf y_j|x)}
\frac{\delta g\left(\frac{\mathbf x - y_j}{r}\right)}{\delta \omega_l}.
\label{eq:diff-conv-mc}
\end{equation}

\mysubsection{Estimating the PDF}{SampleDensityFunction}
Unfortunately, the sample density itself is not given but must be estimated from the samples themselves. To do so, we employ Kernel Density Estimation~\cite{rosenblatt1956,parzen1962}.
The function estimated is high where the samples are dense and low where they are sparse. It is computed  as
\begin{equation}
p(\mathbf y_j|\mathbf x)
\approx
\frac{1}{|\mathcal N(\mathbf x)| \sigma^3}
\sum_{k\in\mathcal N(\mathbf x)}
\left\{
\prod_{d=1}^{3}
h
\left(
\frac{\mathbf y_{j,d}-\mathbf y_{k,d}}\sigma
\right)
\right\},
\label{eq:kde}
\end{equation}

\noindent where $\sigma$ is the bandwidth which determines the smoothing of the resulting sample density function (we use $\sigma=.25 r$), $h$ is the Density Estimation Kernel, a non-negative function whose integral equals $1$ (we use a Gaussian), and $d$ is one of the three dimensions of $\mathbb R^3$.

The PDF of a point $\mathbf y_j$ in respect to a given point $\mathbf x$ is always relative to all other samples in the receptive field. Therefore, density can not be pre-computed for a point $\mathbf y_j$ since its value will be different for each receptive field defined by $\mathbf x$ and radius $r$. Note that in a uniform sampling setting $p$ would be a constant.

\mysection{MC convolution on multiple samplings}{MultiSample}
Our construction does not only allow handling varying sampling densities but also to perform convolution between two (\refSec{TwoSamplings}) or even multiple different samplings (\refSec{MultipleSamplings}).

\myfigure{MultipleSamplings}{Monte Carlo convolution for the same / different sampling \textbf{(horizontal)} and a single / multiple feature channels \textbf{(vertical)}.
Samples are denoted as dots, whereby lines indicate if samplings match. Colors indicate the samplings $\mathcal A,\mathcal B$ and $\mathcal C$.
}

Convolution involving a single sampling is illustrated in \refFig{MultipleSamplings} (a). Here, the sampling $\mathcal A$ of the input and the output is identical. While the previous section has detailed how to account for varying sampling density, we will show in this section how MC convolution can seamlessly handle two or more samplings.

\mysubsection{Two samplings}{TwoSamplings}
A more generalized setting is shown in \refFig{MultipleSamplings} (b). Here, the input sampling is still $\mathcal A$, but the output is on a different sampling $\mathcal C$.

The illustration shows a mapping from a lower sampling to a higher sampling (also called upsampling, transposed convolution or deconvolution). The same principle can be used to reduce the sample resolution (pooling), as necessary for example when an entire point cloud is successively reduced in resolution to produce a single scalar classification value. We will make use of combinations of up- and downsampling in a U-net / encoder-decoder  architecture \cite{Ronneberger2015} for our segmentation application.

Previous work also operated using different samplings in a multi-resolution hierarchy, but using fixed, non-learned interpolation, \eg inverse-distance weighting~\cite{qi2017plusplus} for upsampling operations. Our approach allows learning these mappings instead.

The procedure explained in the previous section simply works in the two-sampling case, as the kernels $g$ are defined on all continuous offset vectors $\mathbf x^{\mathcal C} - \mathbf y^{\mathcal A}$, that can be computed at any output position $\mathbf x^{\mathcal C}$ in sampling $\mathcal C$, and input position $\mathbf y_j^{\mathcal A}$ in sampling $\mathcal A$. Note, that a density estimation has to be performed respectively to $\mathcal A$, the input, as explained in the previous section.

\mysubsection{Multiple samplings}{MultipleSamplings}
Another unique advantage of our construction is to relax the sampling requirements not only between the input and output sampling $\mathcal A$ and $\mathcal C$, but also between the different inputs. The case of multiple input channels is seen in \refFig{MultipleSamplings} (c) and (d). For \refFig{MultipleSamplings} (c) the sampling remains identical ($\mathcal A$) between two input channels and the output. In \refFig{MultipleSamplings} (d) the sampling is mutually different between both inputs and the output.

A typical application of this multiple-sampling approach is to consume information from multiple resolutions in a hierarchy at the same time. Our example shows a learned up-sampling from five samples at $\mathcal A$ and three samples at level $\mathcal B$ which are combined into a ten-sample result $\mathcal C$. Note, that in the multiple-samplings case, density estimation is to be performed relative to $\mathcal A$ when convolving samples from $\mathcal A$ and relative to $\mathcal B$ when processing points from $\mathcal B$.

Combining output from multiple previous layers has also been used in DenseNet~\cite{Huang2016}, but the classic tabulated kernels do not admit to construct dense links between different samplings.

We are not limited to use the same receptive field size on all samplings but each can choose its own, such that the number of samples falling into the receptive field remains roughly constant. Note, that any deviation from this desired constant sample count, as well as variation of density inside the receptive field, is compensated for using the density estimation.

\myfigure{DenseUnstructured}{Features upsampled from different levels for a single black point $\mathbf x$ on layer $i$ with respect to two previous layers.
The receptive field content $f_{i-1}$ and $f_{i-2}$ is shown on the right, with their respective densities $p_{i-1}$ and $p_{i-2}$ in pink.
Each layer is MC-convolved using an individual MLP kernel $g_{i-1}$ and $g_{i-2}$ which results are concatenated to create the feature vector for point $x$ of layer $i$. 
We guarantee the same maximum number of points in all receptive fields by maintaining the same ratio between the Poisson disk and the radius of the convolutions.}

A particular embodiment of multi-sampling MC convolution is shown in~\refFig{DenseUnstructured}, where information from two 2D point clouds with different resolutions are up-sampled into a third one with a higher resolution.
Note how the receptive field grows in the level with a lower resolution.

\mysection{Poisson disk hierarchy}{PoissonDisk}
Deep image processing routinely reduces -- and later increases again -- the image resolution to make use of both local and global information.
The same is achieved in deep point cloud processing~\cite{qi2017} usually using Farthest Point (FP) sampling \cite{eldar1997}. In this work, we favor using Poisson disk (PD) sampling~\cite{Cook1986} instead. This has two reasons: scalability and the ability to preserve the sampling pattern while bounding the sample count per unit measure.

\paragraph{Realization}
PD is realized as a network layer using the Parallel Poisson Disk Sampling algorithm~\cite{Wei2008}. The input for these layers are any point sampling and a parameter $r_\mathrm p$ controlling the PD radius.
The output is a sampling with a minimum distance between points equal to $r_\mathrm p$. Note that this does not bound the distance from above, so areas with distances much larger than $r_\mathrm p$ can remain.

Multiple PD layers can be combined to create a multi-resolution hierarchy. We use this in combination with multi-samplings convolution (\refSec{MultiSample}) to build an encoder-decoder network~\cite{Ronneberger2015} for point clouds.

Note that, contrary to other sampling approaches, this technique generates a non-fixed number of samples. Therefore, our networks cannot take advantage of acceleration techniques commonly used by deep learning frameworks in which the memory required for a forward pass is reserved in advance. However, we still achieve good performance as is presented in Section~\ref{sec:Performance}.

\paragraph{Scalability}
The main practical reason to use PD is scalability to large point clouds as evaluated in~\refTbl{PoissonDiskScalability}. We see, that PD sampling presents a low compute time for the different model sizes (linear to the number of points in the model). FP sampling, on the other hand, does not scale well, requiring more than 100 seconds to sample 100\,k points from a 1,000\,k points model.

\begin{table}[t]
\caption{
Time for sampling a tenth of the points using different algorithms.
}%
\label{tbl:PoissonDiskScalability}%
\begin{tabularx}{0.98\linewidth}{rrrrr}
&
\multicolumn{1}{c}{100/1\,k}&
\multicolumn{1}{c}{1\,k/10\,k}&
\multicolumn{1}{c}{10\,k/100\,k}&
\multicolumn{1}{c}{100\,k/1000\,k}\\
\toprule
\textsc{\small Poisson disk}&10.4\,ms&21.7\,ms&136.4\,ms&1,304.9\,ms\\
\textsc{\small Farthest Point}&2.4\,ms&11.9\,ms&657.1\,ms&108,682.8\,ms\\
\bottomrule
\end{tabularx}
\end{table}

\paragraph{Sample count bound}
Moreover, PD allows us to limit the number of points within the receptive fields of our convolutions. The Kepler conjecture~\cite{hales2017}, provides an upper bound to the number of points inside the receptive fields: This is illustrated in \refFig{ReceptiveFieldPacking}.
Starting from a non-uniform sampling \textbf{(a)}, PD will retain non-uniformity whilst maintaining a minimal distance between points \textbf{(c)}. Therefore, any ball, such as the receptive field of our approach (pink circle) will at most retain a bound number 
\begin{gather}
n < 
\frac
{\pi\left(r + \frac{r_\mathrm p}{2}\right)^3}
{3\sqrt{2}{r_\mathrm p}^3},
\qquad\text{when}\qquad
r_\mathrm p \leqslant r\nonumber,
\label{eq:maxpts}
\end{gather}
of  balls of the Poisson disk radius $r_\mathrm p$ (marked in green) in a receptive field of radius $r$. 

\myfigure{ReceptiveFieldPacking}{
	Result of Farthest Point (FP) (selecting a fix number of samples) and Poisson Disk (PD) sampling for non-uniformly sampled inputs.
    Blue dots are samples, the pink circle denotes the receptive field.
    Grey circles in FP (\textbf b) are samples not fulfilling a minimal distance (overlap).
        Green circles in PD (\textbf c) are balls around samples that can be packed inside the receptive field.
}

In practice, the number of points is much lower than this limit since we learn from points sampled on the surface of a 3D object. As it is illustrated in Figure~\ref{fig:DenseUnstructured}, we can take advantage of this fact to compute features at different scales by maintaining a constant ratio between $r$ and $r_\mathrm p$. We found that a ratio between $4$ and $8$ provides enough samples in our receptive fields ($\sim30$).

The most commonly used implementation of FP sampling~\cite{qi2017} for learning on point clouds, selects a fixed number of samples from the input point cloud. 
Contrary to PD, this method cannot bound the number of points that fall in the receptive fields \textbf{(b)}.
In order to generate a sampling with the same properties as PD, this method would need to be modified to select a variable number of points based on the distance of each new sample to the subset of already selected ones.

\mysection{Implementation}{Implementation}
In this section, we describe several implementation details of the building blocks of the proposed learning approach.

\paragraph{PDF computation}
Our implementation differs from averaging the individual contributions of the neighboring points only by the requirement to divide by their probability values $p$. Therefore, our MC approach requires two steps: computing all values of $p$ and querying them during MC convolution.

To compute $p$, we create a voxel grid with cell size $r$, as in \cite{green2008cuda}, which enables us to perform the desired computation in time and space linear wrt.\ the number of points.
Further scalability would be provided using hashing instead of a regular grid as proposed by Teschner~\etal\shortcite{teschner2003optimized}. 

For lookup performance, the neighbor indices $\mathcal N(\mathbf x)$ of all points $\mathbf x$ are stored in a flat list, which is indexed by a start and end index stored at each grid cell. Additionally, we compute the values of $p(\mathbf y_j|\mathbf x)$ and store them in a list of the same size. Both lists can get arbitrarily long with arbitrarily high density, which decreases performance during the computation of our convolution for large values of $r$. However, maintaining a good ratio between the receptive field and the PD radius provides an upper bound on the number of neighbors of each point.

Based on the thus constructed data structures, the neighbor indices, as well as the precomputed PDF, can now be looked-up in constant time using the computed information. When looking up the neighbor values, they are multiplied by the kernel weight $g((\mathbf x-\mathbf y_j)/r)$ and divided by the $p(\mathbf y_j|\mathbf x)$.

Note that the voxel grid in which points are distributed is computed in parallel on the GPU, resulting in different point orderings within the same cell for different executions. This introduces randomness in the output sampled point clouds of our PD sampling strategy, which is preferred during learning. \refSec{Evaluation} evaluates the effect of the randomness on the resulting accuracy.

\paragraph{Multilayer Perceptron Evaluation}
Evaluating the MLP during MC convolutions requires a considerable amount of GPU memory when using implementations as provided by standard frameworks. This limits the number of computed features per point and the number of convolutions used in the network architecture. In order to address these limitations, we have implemented the MLP evaluation in a single GPU kernel. By doing so, our networks do not require to expand the features and coordinates of neighboring points and also do not need to store intermediate MLP results for each layer. However, with this implementation, we are not able to use some features, such as batch normalization, to improve learning.

\paragraph{Batch Processing}
Our layers process point clouds of variable size. Moreover, as described in Section~\ref{sec:MCConvolution}, our convolutions take into account all the neighboring points to carry out their computations. These design choices do not allow us to use the standard tensor approach to process a batch of models in parallel.

In order to support batch processing, we use an extra vector with an integer value for each point, denoting the model identifier to which the point belongs to. We create an acceleration data structure to access neighboring points (as described at the beginning of this section) for each model, and, during the evaluation of the layers, only the appropriate data structure is updated and queried based on the model identifier of each point. This simple approach allows us to process batches of models with variable size in parallel by increasing the memory consumption linearly with the number of points. However, in configurations with reduced GPU memory, a more sophisticated approach can be considered, such as only storing the number of points per model and perform extra computations within the layers.

\mysection{Evaluation}{Evaluation}
Here we report the results obtained when using the machinery explained before in a complete network on specific data for relevant tasks.
To make the reported results comparable, we introduce a dataset with non-uniform sampling in \refSec{TestData} based on current benchmark data.
Next, we describe the specific tasks carried out in \refSec{Tasks} and the methods used in \refSec{Methods}, before reporting actual quantitative evaluation results in \refSec{Results}.
Moreover, we report the results of applying our networks to process real-world datasets in \refSec{RealWorldData}.
And, lastly, we introduce additional experiments in \refSec{Variants} and present the computational efficiency of our networks in \refSec{Performance}.

\myfigure{Nonuniformity}{
Different sampling protocols applied to the same object.
\textsc{Uniform} is uniformly random.
\textsc{Lambertian} depends on the orientation, here shown as an arrow. Locations facing this direction are more likely to contain points.
For \textsc{Gradient}, the likelihood of generating a point decreases along a direction, here shown as an arrow again.
For \textsc{Split}, the shape is split into two halves, here shown as a dotted line, where each part is samples uniformly random, but with different density.
\textsc{Occlusion} also depends on the orientation. Only locations visible from this direction contain points.}

\mysubsection{Non-uniformly sampled test data}{TestData}
Non-uniform test data is found in typical scanned scenes.
However, existing benchmarks provide data sets of uniformly sampled models~\cite{Wu2015, Yi2016}.
In order to evaluate the performance of our and other state-of-the-art networks, we generated our own data set by artificially producing non-uniform samplings (see \refFig{Nonuniformity}).
This allows us to explicitly study the effect of sampling.

To produce such data, we start from a uniformly sampled point cloud, and at each point perform rejection sampling, whereby the rejection probability is computed according to one of five protocols:
\textsc{Uniform} (no rejection),
\textsc{Split} (probability is either a random constant smaller than 1 in a random half-space or 1, we used 0.25 in our tests),
\textsc{Gradient} (probability is proportional to the projection onto the largest bounding box axis), \textsc{Lambertian} (probability is proportional to the clamped dot product between the surface normal and a fixed ``view'' direction) and \textsc{Occlusion} where probability is one except for points invisible from a certain direction.
These sampling protocols are applied during the training and testing phase on the data sets of the different benchmarks.
Thus, each time a model is loaded, we apply one of these sampling techniques (the one we are currently testing) with a random seed to generate the probabilities of each point and the view direction where applicable.

\mycfigure{Alltogether}{
   Comparison of our segmentation result for uniform \textbf{(second row)} and non-uniform samplings \textbf{(third row)} to the ground truth \textbf{(first row)}.
     Non-uniform sampling use the \textsc{Gradient} (first and second columns), \textsc{Lambert} (third and fourth columns), and \textsc{Split} (fifth and sixth columns) protocols.
}

\mysubsection{Tasks}{Tasks}
In this section, we describe the tasks used to evaluate our networks: classification, segmentation, and normal estimation.

\paragraph*{Classification} This task assigns a label to each point cloud as a whole.
Its performance is measured in percentage of correct predictions (more is better).
We used the resampled version of ModelNet~40~\cite{Wu2015} provided by Qi~\etal~\shortcite{qi2017}.
This dataset is composed of 12,311 point clouds uniformly sampled from objects of 40 categories.
The official split is composed of 9,843 models in the train set and 2,468 in the test set.
The models were sampled into $1,024$ points according to the different protocols of \refSec{TestData}

\paragraph*{Segmentation} This task assigns a label to every point.
It can be quantified  by its intersection-over-union (IoU) metric (more is better).
We segment the 16,881  point clouds of ShapeNet~\cite{Yi2016}, uniformly sampled from 16 different classes of objects, each one composed of between $2$ and $6$ parts, making a total of $50$ parts.
We use the standard train/test split for training~\cite{qi2017plusplus}.
The class of the point cloud is assumed to be known for this task and used as input.
We used as input of our networks the complete point clouds, which are in the range of 1,000 to 3,000 points per object.

\paragraph*{Normal estimation} This task computes a continuous orientation at every point.
It is analyzed by the cosine distance (less is better).
Similar to Atzmon~\etal\shortcite{Atzmon2018}, we use ModelNet~40 for evaluation, taking 1,024 points from each model on the standard train/test split.

\begin{table}[t]
\center
\caption{
Performance of different methods, including ours, \textbf{(rows)} for  different tasks \textbf{(columns)} with different measures (please see text).
}%
\label{tbl:UniformResults}
\begin{tabularx}{\linewidth}{R{3.0cm} R{1.2cm} R{1.2cm} R{1.2cm}}
&
\multicolumn{1}{c}{Classify}&
\multicolumn{1}{c}{Segment}&
\multicolumn{1}{c}{Normals}\\
\toprule
$^{1,2}$Su~\etal\shortcite{Su2018}&-&85.1&-\\
$^{1}$Xu~\etal\shortcite{Xu2018}&\textbf{92.4}\,\%&85.3&-\\
$^{1}$Qi~\etal\shortcite{qi2017plusplus}&91.9\,\%&85.1&.47\\
Qi~\etal\shortcite{qi2017}&89.2\,\%&83.7&-\\
Groh~\etal\shortcite{Groh2018}&90.2\,\%&-&-\\
Shen~\etal\shortcite{Shen2017}&90.8\,\%&84.3&-\\
Wang~\etal\shortcite{Wang2018}&92.2\,\%&85.1&-\\
Li~\etal\shortcite{Li2018}&91.7\,\%&\textbf{86.1}&-\\
Atzmon~\etal\shortcite{Atzmon2018}&92.3\,\%&85.1&.19\\
Klokov~\etal\shortcite{klokov2017escape}&91.8\,\%&82.3&-\\
\midrule
MC (Ours)&90.9\,\%&85.9&\textbf{.16}\\
\bottomrule
\multicolumn{4}{c}{\footnotesize{
$^1$ Additional input \qquad 
$^2$One network per class in segmentation tasks}}\\
\end{tabularx}
\end{table}

\mysubsection{Methods}{Methods}
Please see the Appendix \refSec{Architectures} for training and network details.

\paragraph{Architecture}
A different architecture is used for each task, where we compare three variants: 
The first is PointNet++ \cite{qi2017plusplus} (\texttt{PN++}) with multi-scale grouping (MSG) , serving as a baseline.
The second is our own architecture, but without using MC convolution, \ie kernel-weighted averaging inside the receptive field, denoted as \texttt{AVG}.
The last is our architecture using our Monte Carlo convolution (\texttt{MC}).
Additionally, we compare to a range of methods that have reported results for the uniform data we use.

\paragraph{Evaluation}
We study two variants of training.
The first only trains on uniformly sampled data.
The second only on non-uniformly sampled data, \ie a dataset produced using all sampling protocols from \refSec{TestData} expect \textsc{Uniform}.
Then, the trained models are tested on all 5 sampling protocols described in \refSec{TestData}.
To counterbalance the randomness introduced by the point sampling algorithm, all measurements were averaged across five independent executions.

\begin{table*}[t!]%
\caption{Performance comparison of different methods for three different tasks \textbf{(columns)} for different sampling protocols \textbf{(rows)}.
For each task, we separate training on ``uniform'' (left columns) and on ``non-uniform'' (right columns) data, while test is always done according to a different protocol in each row.
}%
\vspace{-.2cm}
\label{tbl:NonUniformResults}
\begin{tabular}{
l
R{.70cm}R{.70cm}R{.70cm}R{.70cm}R{.70cm}
R{.70cm}R{.70cm}R{.70cm}R{.70cm}R{.70cm}
R{.70cm}R{.70cm}R{.70cm}R{.70cm}R{.70cm}
}
&
\multicolumn{5}{c}{Classification}&
\multicolumn{5}{c}{Segmentation}&
\multicolumn{5}{c}{Normal estimation}\\
\cmidrule(lr){2-6}
\cmidrule(lr){7-11}
\cmidrule(lr){12-16}
\multicolumn{1}{c}{Train$\rightarrow$}&
\multicolumn{3}{c}{\textsc{Uniform}}&
\multicolumn{2}{c}{Non-\textsc{Uniform}}&
\multicolumn{3}{c}{\textsc{Uniform}}&
\multicolumn{2}{c}{Non-\textsc{Uniform}}&
\multicolumn{3}{c}{\textsc{Uniform}}&
\multicolumn{2}{c}{Non-\textsc{Uniform}}
\\
\cmidrule(lr){2-4}
\cmidrule(lr){5-6}
\cmidrule(lr){7-9}
\cmidrule(lr){10-11}
\cmidrule(lr){12-14}
\cmidrule(lr){15-16}
\multicolumn{1}{c}{Test\ $\downarrow$}&
\texttt{PN++}&\texttt{AVG}&\texttt{MC}&\texttt{PN++}&\texttt{MC}&
\texttt{PN++}&\texttt{AVG}&\texttt{MC}&\texttt{PN++}&\texttt{MC}&
\texttt{PN++}&\texttt{AVG}&\texttt{MC}&\texttt{PN++}&\texttt{MC}
\\
\toprule
\textsc{Uniform}&
89.1\%&
88.3\%&
\textbf{90.9}\%&
89.6\%&
\textbf{90.1}\%&
83.6&				
85.6&
\textbf{85.9}&
84.4&
\textbf{85.2}&
.469&
.165&
.\textbf{161}&
.623&
\textbf{.377}\\
\textsc{Split}&
84.4\%&
83.3\%&
\textbf{87.6}\%&
89.1\%&
\textbf{90.6}\%&
82.9&				
\textbf{84.9}&
84.6&
84.9&
\textbf{85.7}&
.581&
.220&
\textbf{.204}&
.595&
\textbf{.222}\\
\textsc{Gradient}&
79.7\%&
82.9\%&
\textbf{87.3}\%&
89.3\%&
\textbf{90.6}\%&
81.7&				
83.8&
\textbf{84.0}&
83.9&
\textbf{85.1}&
.616&
.215&
\textbf{.201}&
.589&
\textbf{.220}\\
\textsc{Lambert}&
\textbf{74.6}\%&
70.1\%&
73.0\%&
89.8\%&
\textbf{90.4}\%&
80.7&				
\textbf{83.1}&
82.7&
83.8&
\textbf{84.9}&
1.61&
.743&
\textbf{.716}&
1.33&
\textbf{.221}\\
\textsc{Occlusion}&
\textbf{74.8}\%&
67.9\%&
72.4\%&
89.5\%&
\textbf{90.2}\%&
81.3&				
\textbf{83.4}&
82.7&
84.3&
\textbf{85.5}&
1.49&
.679&
\textbf{.654}&
1.25&
\textbf{.132}\\
\bottomrule
\end{tabular}
\end{table*}

\mysubsection{Results}{Results}
Here we discuss the results for uniform and non-uniform sampling as summarized in \refTbl{UniformResults} and \refTbl{NonUniformResults}, respectively, for every task.

\paragraph*{Classification}
The results are illustrated in the first column of \refTbl{UniformResults}. Our approach generates competitive classification results for uniform sampling, achieving $90.9\,\%$ of accuracy.
More importantly, our networks presented a robust performance on non-uniformly sampled point clouds (see first block of \refTbl{NonUniformResults}).

When training only on uniformly sampled point clouds, our \texttt{MC} method results in better performance than PointNet++ when testing on uniformly and most of the non-uniformly sampled point clouds.
Our method achieved $90.9\,\%$, $87.6\,\%$ and $87.3\,\%$ on the \textsc{Uniform}, \textsc{Split} and \textsc{Gradient} sampling protocols, in contrast to $89.1\,\%$, $84.4\,\%$ and $79.7\,\%$ achieved by PointNet++. 
For the \textsc{Lambert} and \textsc{Occlusion} protocols, both networks presented a similar performance of around $74-72\,\%$.
In this protocol, half the points for each model are missing which makes the task more difficult.

\mycfigure{RealWorldResults}{Semantic segmentation results of our approach \textbf{(bottom row)} on ScanNet \cite{dai2017scannet}  compared with ground truth \textbf{(top row)}.
}

Furthermore, our \texttt{MC} method presented slightly better performance than PointNet++ in all protocols when training only on non-uniformly sampled point clouds, achieving between $90.1\,\%$ and 90.6\,\% on the different protocols whilst PointNet++ achieved a performance between $89.1\,\%$ and $89.8\,\%$.

Lastly, when compared with the \texttt{AVG} network, our MC convolutions obtained better performance on all protocols.
The \texttt{AVG} network had more difficulties to generalize, as compared with our \texttt{MC} network, presenting severe over-fitting.
In order to prevent over-fitting, we trained the \texttt{AVG} network using the data augmentation strategy followed by PointNet++~\cite{qi2017plusplus}.

The variance of the resulting accuracy over several executions \texttt{MC} was $.0367$, indicating that, the observed mean accuracy is significant.

\paragraph*{Segmentation} Results of our method are compared to the ground truth for different models in \refFig{Alltogether}.
The second column of \refTbl{UniformResults} presents the performance achieved by our segmentation network on uniform data.
These are competitive compared to the state-of-the-art methods, only slightly surpassed by PointCNN~\cite{Li2018}.
Our method achieved $85.9$ whilst PointCNN achieved $86.1$.
Please note, that PointCNN is based on $k$-nearest neighbors convolutions, and, as discussed in Section~\refSec{PreviousWork}, is not the best approach to handle non-uniformly sampled point clouds.

We state segmentation performance for non-uniform sampling in the middle block of \refTbl{NonUniformResults}.
When trained only on uniformly sampled point clouds, our \texttt{MC} network obtained better results on all protocols than PointNet++.
Moreover, when trained on only non-uniformly sampled point clouds, although PointNet++ presented a competitive performance, our \texttt{MC} network also obtained better results than PointNet++ in all protocols.

It is also worth noticing that, as in the classification task, PointNet++ obtained better results when trained with non-uniformly sampled point clouds.
That indicates that the proposed sampling protocols can also be used as a data augmentation technique.

When comparing with the \texttt{AVG} network, \texttt{MC} presented higher accuracy on the uniformly sampled protocol (\texttt{MC} $85.9$ vs.\ \texttt{AVG} $85.6$) and on the \textsc{Gradient} protocol (\texttt{MC} $84.0$ vs.\ \texttt{AVG} $83.8$).
However, \texttt{AVG} performed better on the \textsc{Split} (\texttt{MC} $84.6$ vs.\ \texttt{AVG} $84.9$), \textsc{Lambert} (\texttt{MC} $82.7$ vs.\ \texttt{AVG} $83.1$), and \textsc{Occlusion} protocols (\texttt{MC} $82.7$ vs.\ \texttt{AVG} $83.1$).
Nevertheless, the differences between these networks remain small.

\paragraph*{Normal estimation} The results of this task are shown in the last column of \refTbl{UniformResults}.
On uniform data, our network outperformed state of the art methods, achieving a mean cosine distance of $.16$ and improving thus the accuracy of $.19$ reported by Atzmon \etal~\shortcite{Atzmon2018}.
When tested on non-uniform data (see the last block of \refTbl{NonUniformResults}), our \texttt{MC} approach outperforms PointNet++ in all protocols when trained on both uniformly and non-uniformly sampled point clouds. 
The same network with \texttt{AVG} convolutions also obtained a good performance.
However, \texttt{MC} convolutions obtained better results.

\mysubsection{Real-world data}{RealWorldData}


%
We also applied our method to real-world data for a semantic segmentation task from ScanNet \cite{dai2017scannet}.
This dataset is composed of $1045$ scanned rooms for training, $156$ rooms for evaluation, and $312$ rooms for testing.
The task requires to classify each input point into 20 different categories as shown in \refFig{RealWorldResults}.

We report performance as mean per-class voxel accuracy~\cite{dai20183DMV}; a more meaningful measure than the original ScanNet metric (overall voxel accuracy) used on PointNet++.
Our approach achieves 62.5\,\%, in comparison to 60.2\,\% of PointNet++, 50.8\,\% of ScanNet, and 54.4\,\% of the method proposed by Dai and Nie{\ss}ner~\shortcite{dai20183DMV}.
When adding image information from 5 different views, Dai and Nie{\ss}ner achieved 75.0\%.
Our network is able to generate consistent predictions on unannotated points and predict correct classes for incorrect annotated objects on the ground truth (door in the second column and sofas in the third column of \refFig{RealWorldResults}).
These point clouds have up to 600\,k points, that our method can handle in 3$\sim$5 seconds during training without splitting it into chunks.

\begin{figure}[t]
\begin{minipage}{4cm}
\centering
  \begin{tabular}{l}\end{tabular}
  \captionof{table}{Classification accuracy and timing for different MLP sizes.}
  \label{tbl:MLPEval}
  \vspace{-.2cm}
  \begin{tabular}{rrr}
\multicolumn{1}{c}{MLP}&
\multicolumn{1}{c}{Accuracy}&
\multicolumn{1}{c}{Time}\\
\toprule
4&90.8\,\%&21.0\,ms\\
8&90.9\,\%&24.6\,ms\\
16&90.6\,\%&26.2\,ms\\
\bottomrule
\end{tabular}
\end{minipage}%
\ \ \ \ \ 
\begin{minipage}{3.5cm}
\centering
\begin{tabular}{l}\end{tabular}
\captionof{table}{Train (one epoch) and test time for different tasks.}
  \label{tbl:ComputeTime}
  \vspace{-.2cm}
 \begin{tabular}{rrr}&
\multicolumn{1}{c}{Train}&
\multicolumn{1}{c}{Test}\\
\toprule
Classify&454\,s&24.6\,ms\\
Segment&2160\,s&87.4\,ms\\
Normals&426\,s&28.1\,ms\\
\bottomrule
\end{tabular}
\end{minipage}
\end{figure}

\mysubsection{Variants}{Variants}
\paragraph{Poisson disk hierarchy}
In order to evaluate the MC convolutions without considering the Poisson Disk hierarchy, we trained a simple network composed of two convolutions on the normal estimation task.
First we trained it using \texttt{AVG} convolutions, and then using \texttt{MC} convolutions.
%
We found \texttt{AVG} to perform worse than \texttt{MC} for \textsc{uniform} (.305), \textsc{split} (.336) \textsc{gradient} (.334) and \textsc{lambert} protocols (.693).
Similarly, the performance of \texttt{MC} without PD was limited for \textsc{uniform} (.282), \textsc{split} (.312) \textsc{gradient} (.310) and \textsc{lambert} (.662) as well.
This indicates, that at least for normal estimation, using a PD hierarchy is essential.

\paragraph{MLP size}
When comparing the classification accuracy at different MLP sizes (\refTbl{MLPEval}), the maximal accuracy was obtained at $8$ neurons, whilst $4$ neurons achieve better timing.
We decided to use $8$ neurons since it provides the best accuracy-execution time trade-off.

\mysubsection{Computational efficiency}{Performance}

\refTbl{ComputeTime} presents the time required to train an epoch of our networks and the time required to compute a forward pass for a single model.
Training time is measured for one single epoch.
Testing time is the time required to process a forward pass of an individual model.
Due to our parallel implementation of all the algorithms and our acceleration data structures, all networks present competitive performance.
The lowest performance is found for the segmentation network since it requires to compute a high number of features for the initial point set.
All measurements were used an Nvidia GTX 1080 Ti.

\mysection{Limitations}{Limitations}
Besides the many benefits presented in this paper, the proposed approach is also subject to a few limitations.

The main limitation is that we have to rely on KDE to obtain the PDF, which requires to carefully select the bandwidth parameter to obtain a good PDF approximation. 
In the future, we would like to further investigate this shortcoming and inspect more advanced PDF estimations, like the selection of $\sigma$ in KDE using cross-validation, ballooning, or otherwise automate its choice.

While our approach provides an unbiased estimate of the convolution -- assuming the KDE was reliable --  there is a variance-locality trade-off: on one hand, one wants a good locality, small receptive fields and few points in them to allow for fast computation, but these give a noisy estimate. On the other hand, a net with large receptive fields that contain many points is slower to compute, does not localize, but has less noisy estimates.

\mysection{Conclusions}{Conclusions}
We have shown how phrasing convolution as a MC estimate produces results superior to the state-of-the-art in typical learning-based processing of non-uniform point clouds, such as segmentation, classification, and normal estimation.
This was enabled by representing the convolution kernel itself using a multilayer perceptron, by accounting for the sample density function, by using Poisson disk pooling, and by realizing MC up- and down-sampling to preserve the original sample density.
Moreover, the experiments demonstrated that our networks are more robust to over-fitting with better generalization, being able to obtain the best performance without any data augmentation technique (something mandatory in the classification task if the point density is not considered).

Although other methods were able to present competitive performance on some tasks when trained on non-uniformly sampled data, it is not always possible to predict the sampling of our future input data in real-world scenarios.
Our model's ability to generalize and become robust to unseen samplings is of key importance for the success of this type of networks in real-world tasks.

In future work, we would like to apply our idea to inputs of higher dimensionality, such as animated point clouds or point clouds with further attributes, such as color.
Another direction for future research could be to consider what a non-uniform density means when dealing with triangular or tetrahedral meshes, that typically do not come with uniform sampling. 


\begin{acks}
We would like to thank the reviewers for their comments. This work was partially funded by the \grantsponsor{1}{Deutsche Forschungsgemeinschaft }(DFG) under grant \grantnum{1}{RO 3408/2-1} (ProLint), the \grantsponsor{2}{Federal Minister for Economic Affairs and Energy } (BMWi) under grant \grantnum{2}{ZF4483101ED7} (VRReconstruct), \grantnum{3}{TIN2017-88515-C2-1-R } (GEN3DLIVE) from the \grantsponsor{3}{Spanish Ministerio de Econom{\'\i}a y Competitividad } by 839 FEDER (EU) funds, and a Google Faculty Research Award. We also acknowledge NVIDIA Corporation for donating a Quadro P6000.
\end{acks}

\bibliographystyle{ACM-Reference-Format}
\bibliography{biblio}

\begin{appendix}

\mysection{Architectures}{Architectures}

\mycfigure{Networks}{
	Network architectures used for the classification, segmentation, normal estimation, and semantic segmentation tasks.
    The three different building blocks used to generate our networks are described at the bottom of the figure:
    \emph{$1 \times 1$ convolutions}, which reduce or increase the number of point features by combining them;
    \emph{Spatial convolutions}, which use a level of the point hierarchy as the center of the convolution and another level to sample the feature functions in order to generate a set of new point features; 
    And \emph{multi-layer perceptrons} (MLP), which are composed of three fully-connected layers.
} 

\mysubsection{Classification}{ClassificationArchitecture}
Our classification architecture is composed of several levels (\refFig{Networks}).
Each computes a convolution on a point cloud, uses Poisson disk sampling to reduce the number of points, and performs a down-sampling operation to compute the features of the new points.

The convolution of the first level is a multi-feature convolution in order to increase the number of features. 
However, in deeper levels, we use a single-feature convolution for performance considerations.
In order to incorporate combinations of features in such levels, we introduce $1 \times 1$ convolutions between the spatial convolutions.

Before each layer, we add a batch normalization layer to improve training, and a ReLU layer to introduce non-linearities.
Moreover, similar to Dense blocks \cite{Huang2016}, we incorporate skip links between the output of the down-sampling layers and the output of the spatial convolution.

The classification network generates a point cloud hierarchy of four levels by iteratively using Poisson Disk sampling on the input point cloud with radius $.1$, $.4$, and $\sqrt{3}$ (we use the original point cloud as the first level).
The last level ($4$) is composed of only a single point since the Poisson Disk radius used to compute it was equal to the diagonal of the bounding box.

The final output of the network is a feature vector describing the model which is processed by an MLP with two hidden layers.

In order to increase the robustness of the classification under poor samplings, this architecture was replicated in two different paths, which generate different probability vectors that are added together to create the final class probabilities.
However, the second path is composed of only two layers and works directly with the second level of the point cloud hierarchy.


\paragraph{Training}
We use cross-entropy loss with an Adam optimizer, a batch size of $32$, and an initial learning rate of $.005$.
The learning rate is divided by half after every $20$ epoch.
To prevent over-fitting we used a drop-out probability of $.5$ in the final MLPs and a drop-out probability of $.2$ for the point features before each layer.
Moreover, we selected a point from the dataset with a probability of $.95$, which varied the input points during training.
In order to obtain a network robust to models with different samplings, during training, we deactivate one or none of the two paths of the network.
The network was trained for $200$ epochs.

\mysubsection{Segmentation}{SegmentationArchitecture}
This network computes a four-level point hierarchy by iteratively applying our Poisson disk sampling algorithm with radius $.025$, $.1$, and $.4$.
It makes use of an encoder-decoder architecture (\refFig{Networks}).


Since the class of the model is assumed as input, as in PointNet++~\cite{qi2017plusplus}, we concatenate a \emph{one-hot} vector containing this information with the output of our network.
This information is processed by an MLP composed of two hidden layers with $512$ and $256$ neurons, and $50$ outputs, which generates the parts probabilities.

\paragraph{Training}
We trained using a cross-entropy loss with an Adam optimizer and a batch size of $32$.
We used an initial learning rate of $.005$ which was scaled by $.2$ every $20$ epochs.
As in the classification networks, we used a drop out rate of $.5$ in the final MLP and $.2$ before each layer.
Since in this task we used the complete point set as input, we used a probability of $.2$ to drop out a point during training.
We trained our network for $90$ epochs.

\mysubsection{Normal estimation}{NormalEstimationArchitecture}
Our network has an encoder-decoder architecture which generates a point hierarchy of three levels by iteratively applying our Poisson Disk Sampling algorithm with radius $.1$ and $.4$.



\paragraph{Training}
We used a cosine distance loss using an Adam optimizer and a batch size of $16$.
Initially, we used a learning rate of $.005$ which we decreased by half every $20$ epochs.
We trained our network for $160$ epochs.

\mysubsection{Semantic segmentation}{SemanticSegmentationArchitecture}
For the semantic segmentation task on real-world datasets, we use a similar architecture to the one used for the segmentation task.
However, since the datasets used in this task are complete rooms of varying size, this network architecture has some differences with the segmentation network  (see \refFig{Networks}).
The most important difference is that, due to the different sizes of the rooms, we define the radius of our operations in meters, in contrast to the segmentation network in which were defined relative to the bounding box.
Moreover, since the point clouds for this task are composed of a higher number of points, the network computes a point hierarchy of 5 levels instead of 4 by applying Poisson Disk sampling with radius .1, .2, .4, and .8 meters.
Lastly, in order to reduce the number of operations and the memory consumption, we do not compute a convolution in the first level of the hierarchy.
Instead, we use a pooling operation to compute features in the second level based on the input point cloud.

\paragraph{Training}
We trained using a cross-entropy loss with an Adam optimizer.
As in the segmentation task, we used an initial learning rate of $.005$ which was scaled by $.2$ every $20$ epochs.
In order to prevent over-fitting, we used a weight decay factor of $0.0001$ and a drop out rate of $.5$ in the final MLP and $.2$ before each layer.
Moreover, we used a probability of $.15$ to drop-out a point during training.
We trained our network for $100$ epochs.

Since the models on the dataset have a varying number of points, in the range of [10\,k, 550\,k], instead of defining a fixed number of rooms per batch we defined a fixed number of points per batch.
Each train step, we select as many rooms as possible until we fill the budget of 600\,k points.

Furthermore, the number of points per class is not equally distributed (most of the points belong to the classes \texttt{floor} and \texttt{wall}).
In order to train a model which is able to classify points for all the classes and not only the most common, we weighted the losses of each individual points based on the class.
Moreover, in contrast to previous approaches, we consider all the unannotated points in our input point clouds.
However, the loss generated by these points are weighted by 0 and they are not considered during the computation of the performance metric.

\end{appendix}
\end{document}